\documentclass{article} 
\usepackage{iclr2024_conference,times}

\usepackage{amsmath,amsfonts,bm}

\def\eqref#1{equation~\ref{#1}}

\def\1{\bm{1}}

\DeclareMathAlphabet{\mathsfit}{\encodingdefault}{\sfdefault}{m}{sl}
\SetMathAlphabet{\mathsfit}{bold}{\encodingdefault}{\sfdefault}{bx}{n}

\usepackage{hyperref}
\usepackage{url}
\usepackage{graphicx}
\usepackage{textcomp}

\title{Predicting species occurrence patterns from partial observations}

\author{Hager Radi Abdelwahed \\
Mila: Quebec AI Institute\\
\texttt{hager.radi@mila.quebec}
\And
Mélisande Teng\\
 Mila, Université de Montréal\\
  \texttt{tengmeli@mila.quebec}
\And
David Rolnick \\
Mila, McGill University\\
\texttt{drolnick@cs.mcgill.ca}
\\
}

\iclrfinalcopy 
\begin{document}

\maketitle

\begin{abstract}
To address the interlinked biodiversity and climate crises, we need an understanding of where species occur and how these patterns are changing. However, observational data on most species remains very limited, and the amount of data available varies greatly between taxonomic groups.  We introduce the problem of predicting species occurrence patterns given (a) satellite imagery, and (b) known information on the occurrence of other species. To evaluate algorithms on this task, we introduce SatButterfly, a dataset of satellite images, environmental data and observational data for butterflies, which is designed to pair with the existing SatBird dataset of bird observational data. To address this task, we propose a general model, R-Tran, for predicting species occurrence patterns that enables the use of partial observational data wherever found. We find that R-Tran outperforms other methods in predicting species encounter rates with partial information both within a taxon (birds) and across taxa (birds and butterflies). Our approach opens new perspectives to leveraging insights from species with abundant data to other species with scarce data, by modelling the ecosystems in which they co-occur.

\end{abstract}

\section{Introduction}

The interconnectedness of the climate and biodiversity crises has been widely acknowledged~\citep{portner2021ipbes}. Biodiversity and associated ecosystem services play a crucial role in both mitigating and adapting to climate change, as well as being severely threatened by it. It is essential to understand species distributions to inform land use decisions and adaptation measures. Using machine learning and remote sensing data has proved promising for a variety of tasks in biodiversity monitoring~\citep{wang2010remote, reddy2021remote} including species distribution modelling ~\citep{beery2021species,estopinan2022deep, joly2022overview}, improving on traditional methods using only environmental data. Moreover, the integration of machine learning and citizen science has proved helpful to monitor biodiversity at scale, automating labelling and decreasing the need for costly field surveys~\citep{lotfian2021partnership, antonelli2023integrating}.

Recently, the SatBird \cite{teng2023satbird} dataset was proposed for the task of predicting bird species encounter rates from remote sensing imagery, leveraging observational data from the citizen science database eBird~\citep{eBird:HCLN}. While this framework is useful to model species' distributions in places where no data has been recorded before, birds represent an atypical case since relatively large amounts of data are available. For most taxonomic groups, records are present in significantly fewer locations. For example, the eButterfly database~\citep{prudic2017ebutterfly} modelled after eBird provides high quality presence-absence data for butterfly species but has smaller geographical coverage and less observation reports than eBird. Given this challenge, a potentially promising approach is to leverage the relationships between the occurrence patterns of different species, which are used extensively by ecologists e.g.~in joint species distribution models. We design a task to explore this direction, using extensive bird occurrence data to help learn from sparser butterfly occurrence data, given the correlation in abundances between these two taxonomic groups~\citep{gilbert1975butterfly, debinski2006quantifying, eglington2015patterns}. Butterflies are particularly affected by climate change, with many species being adapted to specific environmental conditions~\citep{hill2002responses, rodder2021climate, alvarez2024heterogeneity}. We hope furthermore that such techniques will be useful in generalizing ML predictions to other under-observed but hyper-diverse taxa threatened by climate change, such as amphibians, freshwater fish, and plants.

In this paper, we consider the practical setting in which we want to do checklist completion for species encounter rates. Our main contributions are as follows:
\begin{itemize}
\setlength\itemsep{0em}
    \item We introduce SatButterfly, a dataset for predicting butterfly species encounter rates from remote sensing and environmental data. Importantly, a subset of the data are colocated with existing data from SatBird, making it possible for the first time to leverage cross-taxon relationships in predicting species occurrence patterns from satellite images.
    \item We propose R-Tran, a model to train and predict species encounter rates from partial information about other species. R-Tran uses a novel transformer encoder architecture to model interactions between features of satellite imagery and labels.
    \item We evaluate R-Tran and other methods in predicting species encounter rates from satellite images and partial information on the encounter rates of other species. We find that R-Tran surpasses the baselines while providing flexibility to use with a variable set of information. 
\end{itemize}

\paragraph{Problem definition} We consider the regression task of predicting species encounter rates from satellite imagery given partial information.
Our work is in line with frameworks that leverage partial label annotations during inference such as Feedback-prop \citep{feedbackprop_CVPR_2018}, a method designed to be used at inference time given partial labels and iteratively updates the scores of unknown labels using the known labels on the test set only. C-Tran~\citep{cTran} also proposed a transformer-based architecture for data completion, using embeddings from image features, labels and states in a multi-label classification setting. We consider our task in two different settings: (1) within a single taxon (SatBird-USA-summer) where we split bird species into songbirds and non-songbirds and must predict one set of species from the other, and, (2) across taxa (SatBird-USA-summer and SatButterfly), where information for one taxon is given and the other must be inferred. Task 2 reflects the standard imbalance in data between birds and insects, and tests the ability of ML models to leverage abundant bird data to predict butterfly data. Task 1 is designed so as to further investigate the task, with songbirds (Passeriformes) representing a discrete taxon of birds with approximately half the species of the whole. \\
We refer to SatBird-USA-summer as SatBird throughout the rest of the paper.

\section{SatButterfly dataset} \label{sec:satbutterfly_dataset}
We collect and publish SatButterfly, a dataset for the task of predicting butterfly species encounter rates from environmental data and satellite imagery. The dataset consists of remote sensing images and environmental data along with labels derived from eButterfly \citep{prudic2017ebutterfly} observation reports, in the continental USA. Observations in eButterfly are particularly skewed towards North America, and we work specifically with the continental USA to match SatBird. Unlike in SatBird, we consider observations over the full year as most butterflies are non-migratory. We follow the process of data collection, preparation, and splitting as in SatBird, collecting satellite images, environmental data and encounter rates from recorded observations. We propose two versions of the dataset, a) SatButterfly-v1, where hotspots do not overlap with those of SatBird and b) SatButterfly-v2, where observations are collocated with those of SatBird \citep{teng2023satbird}. Each version of the dataset contains $\sim$7,000 data samples.

For the targets, we use presence-absence data in the form of complete checklists from the eButterfly citizen science platform. We extract checklists recorded from $2010$ to $2023$ in the continental USA for a total of $601$ species. We compute the encounter rates for each location $h$ as: $\mathbf{y^h} = (y_{s_1}^h, ..., y_{s_n}^h)$, where $y_s^h$ is the number of complete checklists reporting species $s$ at $h$ divided by the total number of complete checklists at $h$. We aggregate the species observations over $13$ years to construct the final targets. We follow the GeoLifeCLEF 2020 dataset~\citep{cole2020geolifedata} and SatBird in extracting satellite data and environmental variables. Further details are in Appendix \ref{appendix_A}.

\begin{figure}[h]
\begin{center}
\includegraphics[height=9cm]{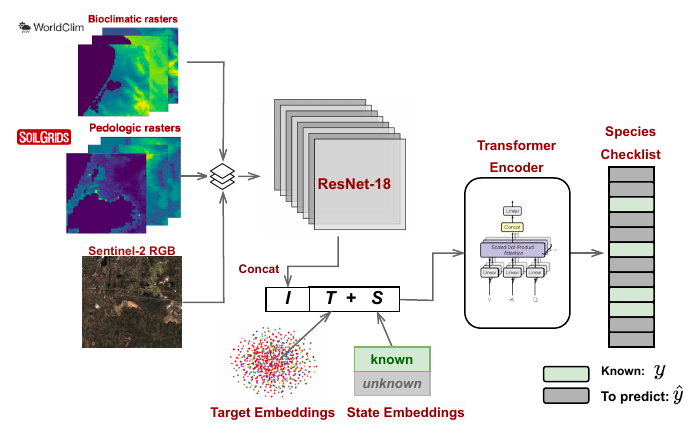}
\end{center}
\caption{R-Tran architecture for predicting species encounter rates with partial information.}
\label{fig:architecture_rtran}
\end{figure}

\section{Methodology: R-Tran}
Our goal is to explore methods for predicting species encounter rates given partial information. We draw inspiration from C-Tran \citep{cTran}, an algorithm that predicts a set of target labels given an input set of known labels, and visual features from a convolutional neural network. A key ingredient of the method is a label mask training objective that encodes the three possible states of the labels (positive, negative, or unknown) during training. We propose the Regression Transformer (R-Tran) ---a model adapted to our setting, which is regression not classification. Figure \ref{fig:architecture_rtran} shows the architecture for R-Tran, which acts as a general framework for predicting species encounter rates given (a) satellite images and environmental data, and (b) observations for a subset of species. Key components of R-Tran are the target embeddings $T$, where we train an embedding layer to represent all possible species classes, and state embeddings $S$, that represent state of each species whether known or unknown through training another embedding layer. Both embeddings are added, and concatenated with extracted features $I$ from satellite and environmental data, then fed into a transformer encoder to model interactions between target classes and features. The model is flexible to accommodate any available partial species information to the full set of species. If no partial information are present, the model can still make predictions from satellite imagery and environmental variables. During training, we provide the full set of available encounter rates and randomly mask out a percentage of these as unknown. This percentage is chosen at random between $0.25n$ and $n$, where $n$ is the total number of targets. During inference, we can provide information on $0$ or more species to predict occurrences of remaining species.

\section{Experiments}
We compare R-Tran to ResNet18 \citep{he2016deep} and Feedback-prop \citep{feedbackprop_CVPR_2018} models. Feedback-prop uses a trained model (in our case, ResNet18 model) and adapts the inference for unknown labels in a way that benefits from partial known labels. More experimental details are available in Appendix \ref{appendix_B}. We evaluate our proposed model, R-Tran, in two settings: a) within taxon: SatBird, b) across taxa: SatBird \& SatButterfly. For SatBird, we use only SatBird-USA-summer, since butterflies are rarely active in the winter.  

\textbf{Within taxon (SatBird)}: we split SatBird's $670$ species into songbirds, denoted set \textit{A}, and non-songbirds, set \textit{B}, composed of $298$ and $372$ species respectively. We train one ResNet18 model on all $670$ species, as well as training individual ResNet18 models on \textit{A} \& \textit{B} species subsets, denoted ResNet18-\textit{A} and ResNet18-\textit{B} respectively. R-Tran is also trained on the full set of bird species.

Table \ref{songbirdvs.nonsongbird} shows results on SatBird: all species, subset \textit{A} and subset \textit{B}. R-Tran and Feedback-prop allow the use of available partial information about species, unlike ResNet18. We observe that R-Tran outperforms other baselines in predicting species \textit{A} given information about \textit{B}, as well as predicting \textit{B} given information about \textit{A}. All results reported are the average of $3$ different seeds, where the standard deviation for MSE and MAE is negligible. Also, metrics reported for set \textit{A} or \textit{B} are masked for that set only.

\begin{table}[t]
\caption{Evaluation within a taxon (SatBird): \textit{A} refers to the subset of songbird classes only, \textit{B} refers to the subset of non-songbird classes.}
\label{songbirdvs.nonsongbird}
\begin{center}
\begin{tabular}{l|c|c|c|c|c}
Model                       & \multicolumn{1}{c|}{MAE[$1^{-2}$]} & \multicolumn{1}{c|}{MSE[$1^{-2}$]}  & \multicolumn{1}{c|}{Top-10 \%} & \multicolumn{1}{c|}{Top-30 \%} & \multicolumn{1}{c}{Top-k \%} \\ \hline
\hline
\multicolumn{6}{c}{All bird species}\\
\hline
ResNet18        & 2.19	& 0.65 & $45.28 \pm 0.61$	& $64.62  \pm0.5$ & $66.26  \pm 0.43$ \\
R-Tran   & 2.13	& 0.67 & $44.5 \pm 0.4$ & $64.2 \pm 0.3$ & $66.1 \pm 0.26$ \\
\hline
\multicolumn{6}{c}{Evaluation on classes \textit{A}: Songbirds}\\
\hline
ResNet18         &3.25	& 1.02 & $53.75\pm 0.66$ & 	$78.94\pm 0.42$ & $69.92\pm 0.48$ \\
ResNet18-\textit{A}       & 3.23	& 1.02 & $53.91\pm 0.23$ &	$79.03\pm 0.28$ & $70.1\pm 0.29$ \\
Feedback-prop \( (A | B) \)             & 2.97 & 0.90 & $56.54\pm 0.6$ & $80.71\pm 0.35$ &  $71.72\pm 0.32$ \\
R-Tran \( (A | B) \)   &\textbf{2.46}	& \textbf{0.72}& $\textbf{59.98}\pm 0.16$	&$\textbf{82.3}\pm 0.1$ & $\textbf{73.45}\pm 0.2$ \\
\hline
\multicolumn{6}{c}{Evaluation on classes \textit{B}: non-songbirds}\\
\hline
ResNet18         & 1.35 & 0.37 & $59.97\pm 0.41$ & $86.14\pm 0.32$ & $59.72\pm 0.35$ \\
ResNet18-\textit{B}      & 1.32 & 0.36 & $60.04\pm 0.29$ & $86.09\pm 0.19$ & $59.61\pm 0.16$ \\
Feedback-prop \( (B | A) \)           & 1.22 &0.32 & $63.09\pm 0.67$ & $87.68\pm 0.4$ & $61.94\pm 0.48$ \\
R-Tran \( (B | A) \)   & \textbf{0.99} & \textbf{0.26}& $\textbf{66.25}\pm 0.08$ & $\textbf{89.28}\pm 0.07$ & $\textbf{64.19}\pm 0.11$ \\
\hline 
\end{tabular}
\end{center}
\end{table}

\textbf{Across taxa (SatButterfly \& SatBird)}: In this setting, we use the full dataset SatBird-USA-summer, joined with SatButterfly. In SatButterfly, we select species that have at least $100$ occurrences, resulting in $172$ species of butterflies. When combining SatBird and SatButterfly, there are locations where only bird species are observed, only butterfly species are observed, or both species observed. For locations where not all classes of species are observed, we use a masked loss to mask out absent species; we provide more details in Appendix \ref{appendix_B}.

We train ResNet18 and R-Tran on SatBird and SatButterfly jointly, where we predict ($670 + 172 = 842$) species classes. We evaluate on the test set where both taxa (birds and butterflies) are observed, where metrics reported are masked for set \textit{A}, or set \textit{B}. Top-$k$ refers to the adaptive top-k metric defined in \cite{teng2023satbird}, where for each hotspot $k$ is the number of species with non-zero ground truth encounter rates in the set of predicted species. In Table \ref{satbird_satbutterfly}, we report results for ResNet18, Feedback-prop and RTran given partial observations of species to predict others. We find that R-Tran improves the performance on species \textit{A} (birds) given information about butterflies, and \textit{B} (butterflies) given information about birds.
While we only evaluated the model in settings where the sets of known and unknown labels are fixed, it should be noted that R-Tran allows for more flexibility than Feedback-prop because at inference time, any arbitrary subset of species could be unknown, and R-Tran does not require multiple test examples to be effective. 

\begin{table}[t]
\caption{Evaluation across taxa (SatBird and SatButterfly): \textit{A} refers to bird species, \textit{B} refers to butterflies. Results are evaluated on the subset of locations with both bird and butterfly data available.}
\label{satbird_satbutterfly}
\begin{center}
\begin{tabular}{l|c|c|c|c|c}
Model                       & \multicolumn{1}{c|}{MAE[$1^{-2}$]} & \multicolumn{1}{c|}{MSE[$1^{-2}$]}  & \multicolumn{1}{c|}{Top-10 \%} & \multicolumn{1}{c|}{Top-30 \%} & \multicolumn{1}{c}{Top-k \%} \\ \hline
\hline
\multicolumn{6}{c}{Evaluation on classes \textit{A}: Birds}\\
\hline
ResNet18        & $2.07$ & $0.62$ & $48.02 \pm 0.2$ & $65.9 \pm 0.15$ & $73.17 \pm 0.15$ \\
Feedback-prop \( (A | B) \)               &2.06& 0.62 & $48.11 \pm 0.16$ & $66.18 \pm 0.19$ & $73.24 \pm 0.1$ \\
R-Tran \( (A | B) \)   & \textbf{2.04} & \textbf{0.6} & $\textbf{50.34} \pm 0.3$ & $\textbf{67.24} \pm 0.4$  & $\textbf{73.85} \pm 0.18$ \\
\hline
\multicolumn{6}{c}{Evaluation on classes \textit{B}: Butterflies}\\
\hline
ResNet18     & 3.81 & 1.55 & $52.35 \pm 0.43$ & $83.74 \pm 0.43$ & $35.69 \pm 0.85$ \\
Feedback-prop \( (B | A) \)  & 3.81& \textbf{1.54} & $52.51 \pm 0.2$ & $83.97 \pm 0.37$ & $35.83 \pm 0.95$ \\
R-Tran \( (B | A) \)   & \textbf{3.6} & 1.56 & $\textbf{52.82} \pm 0.78$ &	$\textbf{84.21} \pm 0.63$ & $\textbf{36.31} \pm 0.8$ \\ 
\hline

\hline
\end{tabular}
\end{center}
\end{table}

\section{Conclusion}
We presented SatButterfly, a dataset that maps satellite images and environmental data to butterflies observations. We benchmarked different models for predicting species encounter rates considering partial information, within a taxon (SatBird), and across taxa (SatBird and SatButterfly). We find that leveraging partial information at inference time is more effective within a taxon than across taxa, and that our novel R-Tran model generally outperforms baselines. Our work offers avenues for combining data from different citizen science databases for joint prediction of species encounter rates across taxa, in particular for cases where some species are less systematically surveyed than others. In future work, we hope to extend this approach to presence-only data, which represents the majority of citizen science observations, such as those from the iNaturalist platform.

\bibliography{iclr2024_conference}

\begin{thebibliography}{27}
\providecommand{\natexlab}[1]{#1}
\providecommand{\url}[1]{\texttt{#1}}
\expandafter\ifx\csname urlstyle\endcsname\relax
  \providecommand{\doi}[1]{doi: #1}\else
  \providecommand{\doi}{doi: \begingroup \urlstyle{rm}\Url}\fi

\bibitem[Cen(2018)]{Census_Bureau_USA_bounderies_shape_file}
{Census Bureau of USA}.
\newblock \url{https://www.census.gov/geographies/mapping-files/time-series/geo/carto-boundary-file.html}, 2018.
\newblock Accessed: 2023-06-06.

\bibitem[{\'A}lvarez et~al.(2024){\'A}lvarez, Walker, Mingarro, Ursul, Cancela, Bassett, and Wilson]{alvarez2024heterogeneity}
Hugo~Alejandro {\'A}lvarez, Emma Walker, Mario Mingarro, Guim Ursul, Juan~Pablo Cancela, Lee Bassett, and Robert~J Wilson.
\newblock Heterogeneity in habitat and microclimate delay butterfly community tracking of climate change over an elevation gradient.
\newblock \emph{Biological Conservation}, 289:\penalty0 110389, 2024.

\bibitem[Antonelli et~al.(2023)Antonelli, Dhanjal-Adams, and Silvestro]{antonelli2023integrating}
Alexandre Antonelli, Kiran~L Dhanjal-Adams, and Daniele Silvestro.
\newblock Integrating machine learning, remote sensing and citizen science to create an early warning system for biodiversity.
\newblock \emph{Plants, people, planet}, 5\penalty0 (3):\penalty0 307--316, 2023.

\bibitem[Beery et~al.(2021)Beery, Cole, Parker, Perona, and Winner]{beery2021species}
Sara Beery, Elijah Cole, Joseph Parker, Pietro Perona, and Kevin Winner.
\newblock Species distribution modeling for machine learning practitioners: A review.
\newblock In \emph{ACM SIGCAS conference on computing and sustainable societies}, pp.\  329--348, 2021.

\bibitem[Cole et~al.(2020)Cole, Deneu, Lorieul, Servajean, Botella, Morris, Jojic, Bonnet, and Joly]{cole2020geolifedata}
Elijah Cole, Benjamin Deneu, Titouan Lorieul, Maximilien Servajean, Christophe Botella, Dan Morris, Nebojsa Jojic, Pierre Bonnet, and Alexis Joly.
\newblock The {GeoLifeCLEF} 2020 dataset.
\newblock \emph{Preprint arXiv:2004.04192}, 2020.

\bibitem[Debinski et~al.(2006)Debinski, VanNimwegen, and Jakubauskas]{debinski2006quantifying}
Diane~M Debinski, Ron~E VanNimwegen, and Mark~E Jakubauskas.
\newblock Quantifying relationships between bird and butterfly community shifts and environmental change.
\newblock \emph{Ecological Applications}, 16\penalty0 (1):\penalty0 380--393, 2006.

\bibitem[Eglington et~al.(2015)Eglington, Brereton, Tayleur, Noble, Risely, Roy, and Pearce-Higgins]{eglington2015patterns}
Sarah~M Eglington, Tom~M Brereton, Catherine~M Tayleur, David Noble, Kate Risely, David~B Roy, and James~W Pearce-Higgins.
\newblock Patterns and causes of covariation in bird and butterfly community structure.
\newblock \emph{Landscape ecology}, 30:\penalty0 1461--1472, 2015.

\bibitem[Ester et~al.(1996)Ester, Kriegel, Sander, and Xu]{dbscan}
Martin Ester, Hans-Peter Kriegel, J\"{o}rg Sander, and Xiaowei Xu.
\newblock A density-based algorithm for discovering clusters in large spatial databases with noise.
\newblock In \emph{Proceedings of the Second International Conference on Knowledge Discovery and Data Mining}, KDD'96, pp.\  226–231. AAAI Press, 1996.

\bibitem[Estopinan et~al.(2022)Estopinan, Servajean, Bonnet, Munoz, and Joly]{estopinan2022deep}
Joaquim Estopinan, Maximilien Servajean, Pierre Bonnet, Fran{\c{c}}ois Munoz, and Alexis Joly.
\newblock Deep species distribution modeling from sentinel-2 image time-series: a global scale analysis on the orchid family.
\newblock \emph{Frontiers in Plant Science}, 13:\penalty0 839327, 2022.

\bibitem[Gilbert \& Singer(1975)Gilbert and Singer]{gilbert1975butterfly}
Lawrence~E Gilbert and Michael~C Singer.
\newblock Butterfly ecology.
\newblock \emph{Annual review of ecology and systematics}, 6\penalty0 (1):\penalty0 365--395, 1975.

\bibitem[He et~al.(2016)He, Zhang, Ren, and Sun]{he2016deep}
Kaiming He, Xiangyu Zhang, Shaoqing Ren, and Jian Sun.
\newblock Deep residual learning for image recognition.
\newblock In \emph{Proceedings of the IEEE conference on computer vision and pattern recognition}, pp.\  770--778, 2016.

\bibitem[Hengl et~al.(2017)Hengl, Mendes~de Jesus, Heuvelink, Ruiperez~Gonzalez, Kilibarda, Blagoti{\'c}, Shangguan, Wright, Geng, Bauer-Marschallinger, et~al.]{hengl2017soilgrids250m}
Tomislav Hengl, Jorge Mendes~de Jesus, Gerard~BM Heuvelink, Maria Ruiperez~Gonzalez, Milan Kilibarda, Aleksandar Blagoti{\'c}, Wei Shangguan, Marvin~N Wright, Xiaoyuan Geng, Bernhard Bauer-Marschallinger, et~al.
\newblock Soilgrids250m: Global gridded soil information based on machine learning.
\newblock \emph{PLoS one}, 12\penalty0 (2):\penalty0 e0169748, 2017.

\bibitem[Hijmans \& al.(2005)Hijmans and al.]{oldworldclim}
Hijmans and al.
\newblock Worldclim 1.4 (historical climate conditions).
\newblock \emph{International journal of climatology}, 25, 2005.

\bibitem[Hill et~al.(2002)Hill, Thomas, Fox, Telfer, Willis, Asher, and Huntley]{hill2002responses}
Jane~K Hill, CD~Thomas, Richard Fox, MG~Telfer, SG~Willis, J~Asher, and B~Huntley.
\newblock Responses of butterflies to twentieth century climate warming: implications for future ranges.
\newblock \emph{Proceedings of the Royal Society of London. Series B: Biological Sciences}, 269\penalty0 (1505):\penalty0 2163--2171, 2002.

\bibitem[Joly et~al.(2022)Joly, Go{\"e}au, Kahl, Picek, Lorieul, Cole, Deneu, Servajean, Durso, Glotin, et~al.]{joly2022overview}
Alexis Joly, Herv{\'e} Go{\"e}au, Stefan Kahl, Luk{\'a}{\v{s}} Picek, Titouan Lorieul, Elijah Cole, Benjamin Deneu, Maximilien Servajean, Andrew Durso, Herv{\'e} Glotin, et~al.
\newblock Overview of lifeclef 2022: an evaluation of machine-learning based species identification and species distribution prediction.
\newblock In \emph{International Conference of the Cross-Language Evaluation Forum for European Languages}, pp.\  257--285. Springer, 2022.

\bibitem[Kelling et~al.(2013)Kelling, Gerbracht, Fink, Lagoze, Wong, Yu, Damoulas, and Gomes]{eBird:HCLN}
Steve Kelling, Jeff Gerbracht, Daniel Fink, Carl Lagoze, Weng-Keen Wong, Jun Yu, Theo Damoulas, and Carla Gomes.
\newblock {eBird}: A human/computer learning network for biodiversity conservation and research.
\newblock \emph{AI Magazine}, 34, 03 2013.

\bibitem[Kingma \& Ba(2014)Kingma and Ba]{adam}
Diederik~P Kingma and Jimmy Ba.
\newblock Adam: A method for stochastic optimization.
\newblock \emph{arXiv preprint arXiv:1412.6980}, 2014.

\bibitem[Lanchantin et~al.(2021)Lanchantin, Wang, Ordonez, and Qi]{cTran}
Jack Lanchantin, Tianlu Wang, Vicente Ordonez, and Yanjun Qi.
\newblock General multi-label image classification with transformers.
\newblock In \emph{Proceedings of the IEEE/CVF Conference on Computer Vision and Pattern Recognition (CVPR)}, pp.\  16478--16488, June 2021.

\bibitem[Loshchilov \& Hutter(2019)Loshchilov and Hutter]{adamW}
Ilya Loshchilov and Frank Hutter.
\newblock Decoupled weight decay regularization.
\newblock In \emph{International Conference on Learning Representations}, 2019.
\newblock URL \url{https://openreview.net/forum?id=Bkg6RiCqY7}.

\bibitem[Lotfian et~al.(2021)Lotfian, Ingensand, and Brovelli]{lotfian2021partnership}
Maryam Lotfian, Jens Ingensand, and Maria~Antonia Brovelli.
\newblock The partnership of citizen science and machine learning: benefits, risks, and future challenges for engagement, data collection, and data quality.
\newblock \emph{Sustainability}, 13\penalty0 (14):\penalty0 8087, 2021.

\bibitem[P{\"o}rtner et~al.(2021)P{\"o}rtner, Scholes, Agard, Archer, Arneth, Bai, Barnes, Burrows, Chan, Cheung, et~al.]{portner2021ipbes}
Hans-Otto P{\"o}rtner, Robert~J Scholes, John Agard, Emma Archer, A~Arneth, Xuemei Bai, David Barnes, Michael Burrows, Lena Chan, WL~Cheung, et~al.
\newblock Ipbes-ipcc co-sponsored workshop report on biodiversity and climate change.
\newblock \emph{IPBES and IPCC}, 10, 2021.

\bibitem[Prudic et~al.(2017)Prudic, McFarland, Oliver, Hutchinson, Long, Kerr, and Larriv{\'e}e]{prudic2017ebutterfly}
Kathleen~L Prudic, Kent~P McFarland, Jeffrey~C Oliver, Rebecca~A Hutchinson, Elizabeth~C Long, Jeremy~T Kerr, and Maxim Larriv{\'e}e.
\newblock ebutterfly: leveraging massive online citizen science for butterfly conservation.
\newblock \emph{Insects}, 8\penalty0 (2):\penalty0 53, 2017.

\bibitem[Reddy(2021)]{reddy2021remote}
C~Sudhakar Reddy.
\newblock Remote sensing of biodiversity: what to measure and monitor from space to species?
\newblock \emph{Biodiversity and Conservation}, 30\penalty0 (10):\penalty0 2617--2631, 2021.

\bibitem[R{\"o}dder et~al.(2021)R{\"o}dder, Schmitt, Gros, Ulrich, and Habel]{rodder2021climate}
Dennis R{\"o}dder, Thomas Schmitt, Patrick Gros, Werner Ulrich, and Jan~Christian Habel.
\newblock Climate change drives mountain butterflies towards the summits.
\newblock \emph{Scientific Reports}, 11\penalty0 (1):\penalty0 14382, 2021.

\bibitem[Teng et~al.(2023)Teng, Elmustafa, Akera, Bengio, Radi, Larochelle, and Rolnick]{teng2023satbird}
M{\'e}lisande Teng, Amna Elmustafa, Benjamin Akera, Yoshua Bengio, Hager Radi, Hugo Larochelle, and David Rolnick.
\newblock Satbird: a dataset for bird species distribution modeling using remote sensing and citizen science data.
\newblock In \emph{Thirty-seventh Conference on Neural Information Processing Systems Datasets and Benchmarks Track}, 2023.
\newblock URL \url{https://openreview.net/forum?id=Vn5qZGxGj3}.

\bibitem[Wang et~al.(2010)Wang, Franklin, Guo, and Cattet]{wang2010remote}
Kai Wang, Steven~E Franklin, Xulin Guo, and Marc Cattet.
\newblock Remote sensing of ecology, biodiversity and conservation: a review from the perspective of remote sensing specialists.
\newblock \emph{Sensors}, 10\penalty0 (11):\penalty0 9647--9667, 2010.

\bibitem[Wang et~al.(2018)Wang, Yamaguchi, and Ordonez]{feedbackprop_CVPR_2018}
Tianlu Wang, Kota Yamaguchi, and Vicente Ordonez.
\newblock Feedback-prop: Convolutional neural network inference under partial evidence.
\newblock In \emph{IEEE Conference on Computer Vision and Pattern Recognition (CVPR)}, June 2018.

\end{thebibliography}
\bibliographystyle{iclr2024_conference}

\newpage
\appendix
\section{Appendix A: Dataset} \label{appendix_A}

In this section, we provide more details about SatButterfly dataset preparation. For the final hotspots, we excluded locations in the sea using the $5$-meter resolution US nation cartographic boundaries provided by the Census Bureau’s MAF/TIGER \cite{Census_Bureau_USA_bounderies_shape_file}. We collected satellite images and environmental data for each location where butterfly species are observed. For Sentinel-2, we collected $10$-meter resolution RGB and NIR reflectance data and RGB true color images, focusing on a $5$ km$^2$ area centered on each hotspot. Our criteria for image selection included a maximum cloud cover of $10$\%, opting for the clearest image within specific time frames: January 1 - December 31, 2022. We assigned one image to each hotspot, based on the assumption that more recent satellite images better reflect our species data, considering the higher frequency of checklists in recent years as compared to earlier periods.

From WorldClim 1.4 \citep{oldworldclim}, we retrieved $19$ bioclimatic variables as rasters with a size of $50\times 50$ and a spatial resolution of approximately $1$ km, centered on each hotspot. These variables include annual trends in temperature, precipitation, solar radiation, wind speed, and water vapor pressure. In addition, we obtained $8$ soil-related (pedologic) variables from SoilGrids \citep{hengl2017soilgrids250m} at a finer resolution of $250$ meters. SoilGrids offers global maps of various soil properties, such as pH, organic carbon content, and stocks, which are generated using machine learning techniques trained on soil profile data and environmental covariates from remote sensing sources.

\begin{table}[t]
\begin{center}
\caption{Number of samples in each split for the two versions of SatButterfly.}
\vspace{2em}
\label{tab:num-samples-splits}
\begin{tabular}{l|cc}

  & SatButterfly-v1 & SatButterfly-v2   \\ \hline
train & $5316$  & $4677$ \\
validation   & $1147$  & $1002$ \\
test  & $1145$  & $1005$ \\
\hline
\end{tabular}
\end{center}
\end{table}

As mentioned earlier, SatButterfly-v2 data is designed to share locations with SatBird, for both birds and butterflies data to be observed. To prepare the data, we perform BallTree-based clustering where SatBird locations are used as centriods. We then search within $1$ km for neighbour butterfly observations using haversine distance. Finally, butterfly targets are aggregated and recorded for SatBird hotspots wherever available. We end up with available butterfly targets in a small subset of SatBird's train/validation/test splits.

\begin{figure}
\includegraphics[width=\textwidth]{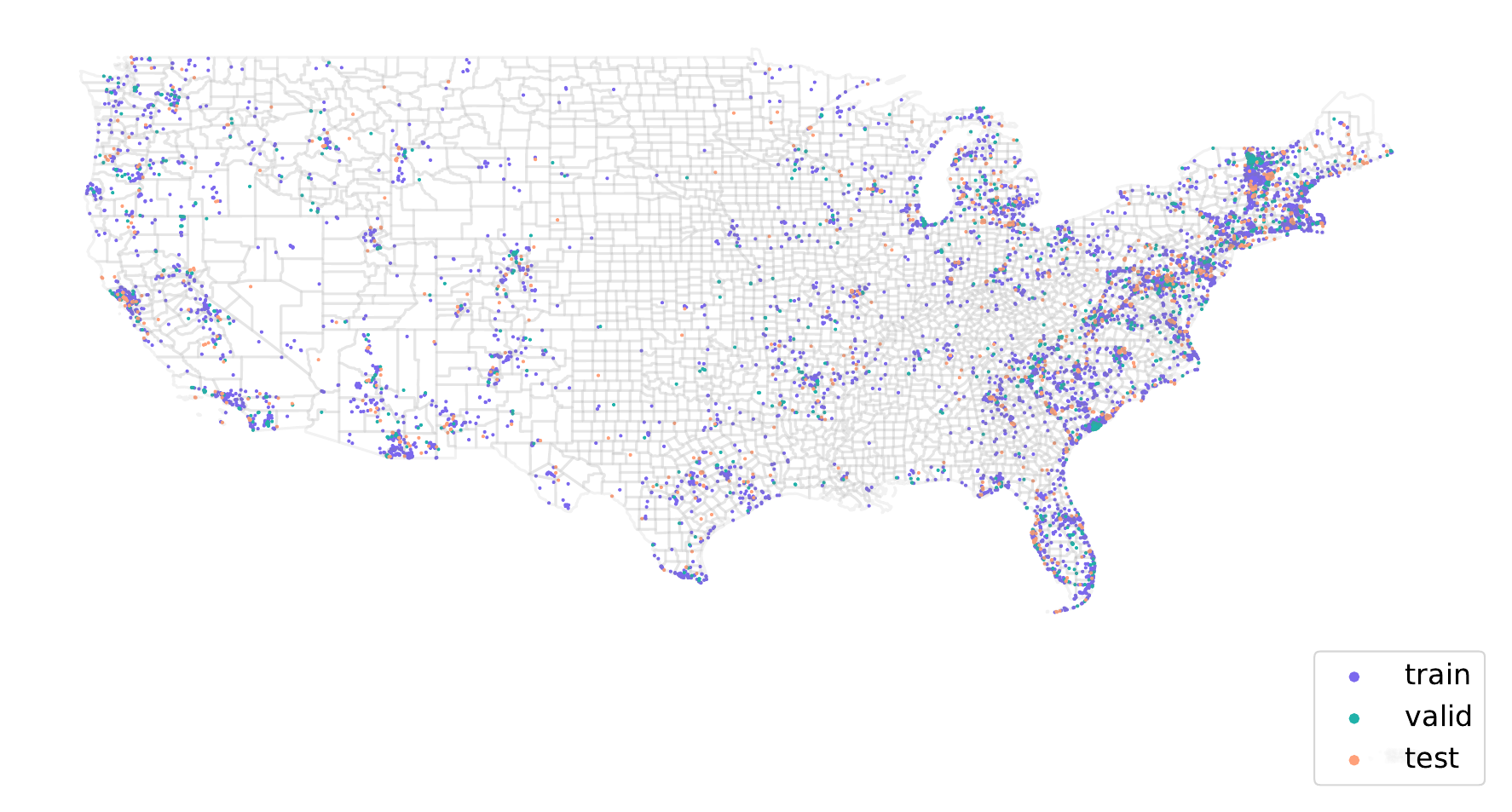}
\label{fig:satbutterfly_v1_datadist}
\caption{Distribution of hotspots across the training, validation, and test sets for SatButterfly-v1}
\end{figure}

\begin{figure}
\includegraphics[width=\textwidth]{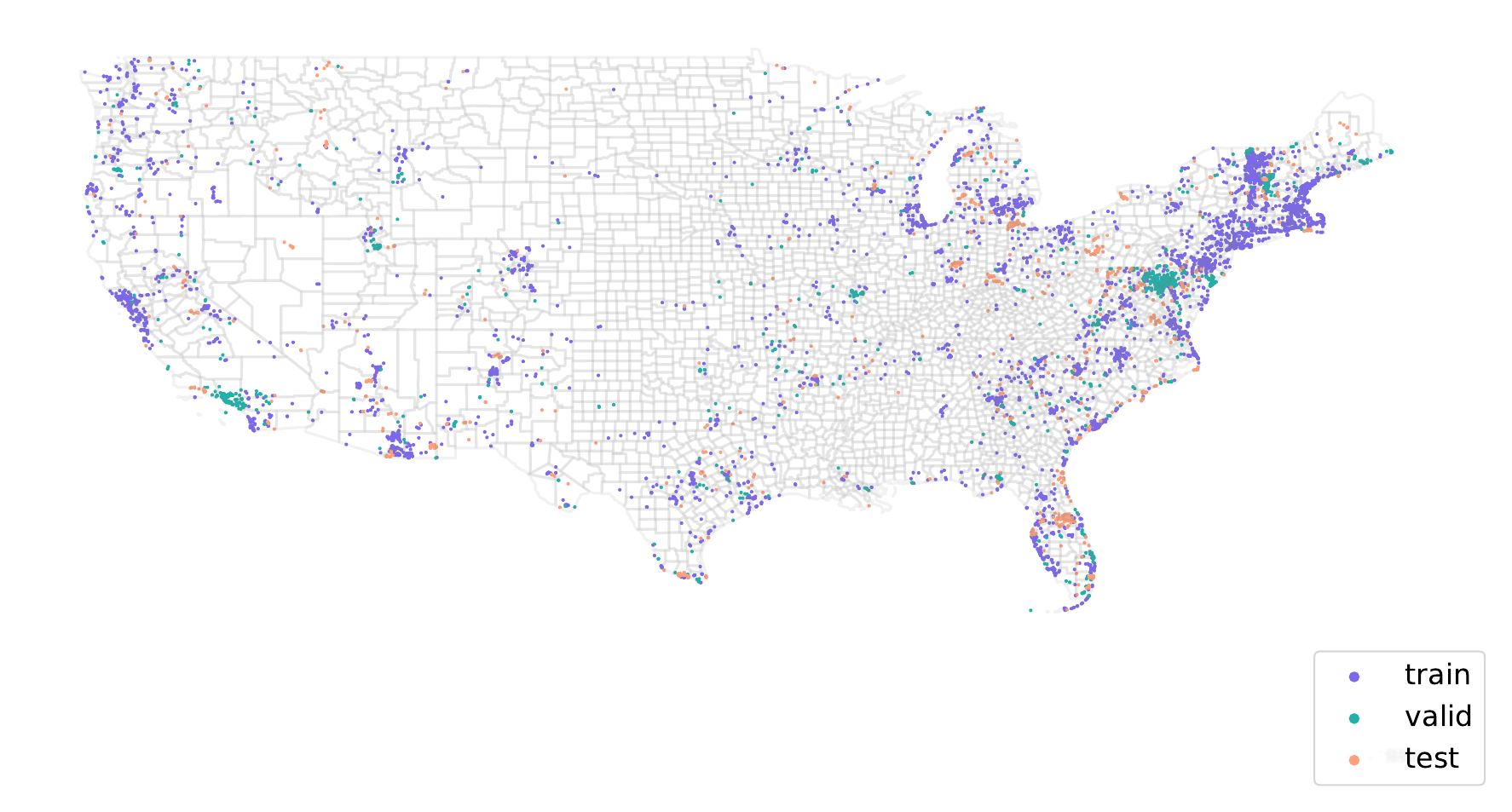}
\label{fig:satbutterfly_v2_datadist}
\caption{Distribution of hotspots across the training, validation, and test sets for SatButterfly-v2}
\end{figure}

Table \ref{tab:num-samples-splits} describes the number of hotspots present in each split for the two versions of SatButterfly dataset. Dataset is split similarily to SatBird using \verb|scikit-learn| DBSCAN \citep{dbscan} clustering algorithm. Figures \ref{fig:satbutterfly_v1_datadist} and \ref{fig:satbutterfly_v2_datadist} shows distribution of data samples in train/val/test splits.

SatButterfly dataset is publicly available through this \href{https://drive.google.com/drive/folders/1dZZqDSTpiL0RMmx9YSc2JAm_t8poqyVc?usp=sharing}{link}.


\section{Appendix B} \label{appendix_B}

\subsection{Experimental Details}
For all models, we use the cross entropy loss:
$\mathcal{L}_{CE}= \frac{1}{N_h} \sum_{h} \mathcal{L}_h = \frac{1}{N_h} \sum_{h} \sum_{s \text{(species)}} -y^s_h \log(\hat{y}^s_h) - (1-y^s_h)\log(1-\hat{y}^s_h) $. For locations where not all classes of species are observed (when combining SatBird \& SatButterfly), we define a masked loss. If we have the loss \( L(y, \hat{y}) \), where \( y \) is the true value and \( \hat{y} \) is the predicted value, and a mask \( M \) (a binary vector or tensor of the same shape as \( y \) and \( \hat{y} \)), the masked loss function is defined as \( L_{\text{masked}}(y, \hat{y}, M) = \frac{\sum_{i} M_i \cdot L(y_i, \hat{y}_i)}{\sum_{i} M_i} \), where \( M_i \) is the \( i \)-th element of the mask \( M \). The loss is calculated only for the unmasked elements where \( M_i = 1 \), and is averaged over the number of unmasked elements.

All models use same input, RGB+ENV, where we concatenate environmental data (bioclimatic, pedologic) and RGB satellite images into different channels resulting in $30$ channels. This is reported in SatBird \citep{teng2023satbird} to be the best performing compared to using RGBNIR. We consider a region of interest of 640 m$^2$, center-cropping the satellite patches to size $64\times64$ around the hotspot and normalizing the bands with our training set statistics. ResNet18 is adapted to take $30$ channels as inputs instead of $3$ channels. ResNet18 and R-Tran are trained for $50$ epochs using Adam \citep{adam} and AdamW \citep{adamW} optimizers respectively. 

\begin{figure}
\centering
\includegraphics[width=\textwidth]{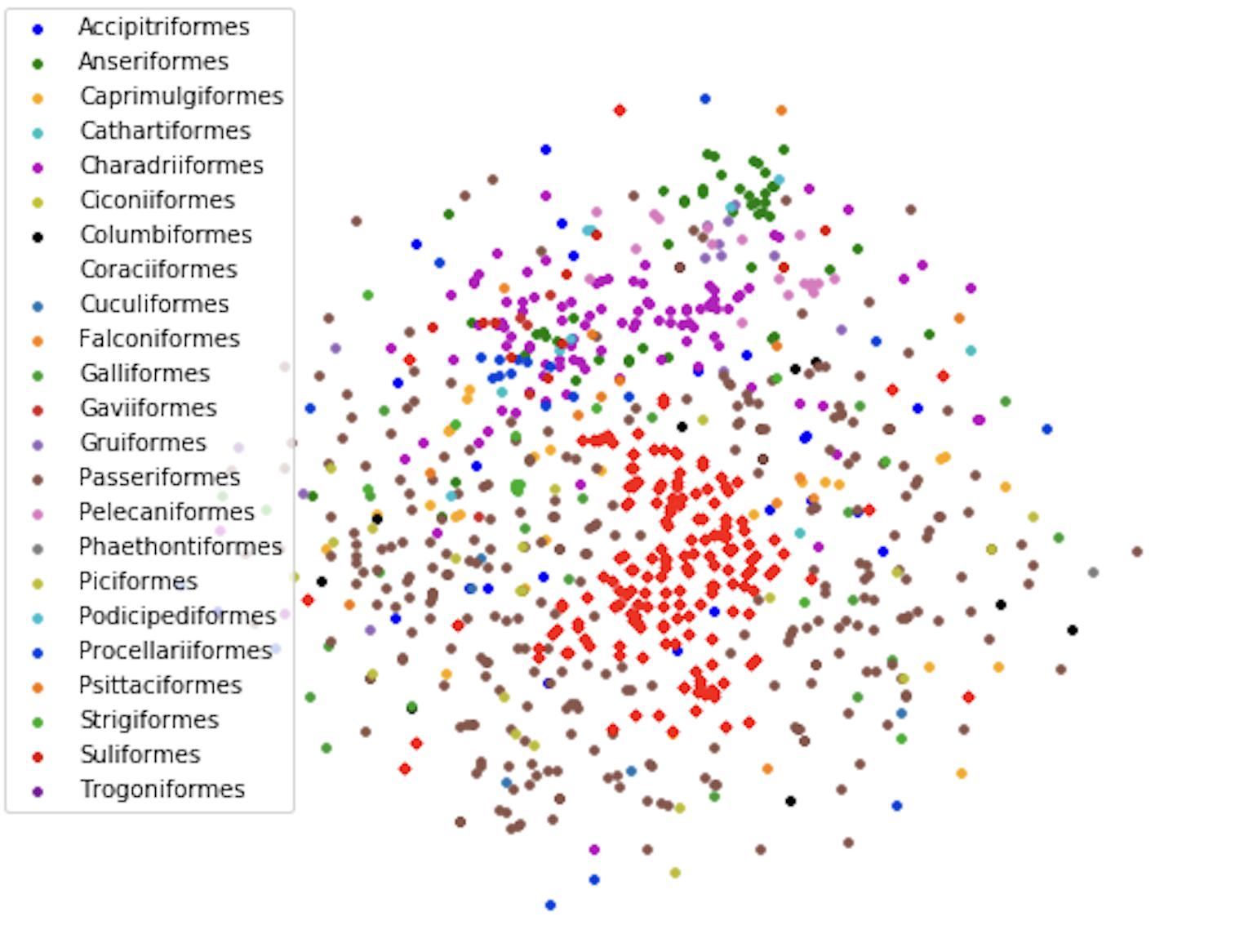}

       \caption{Visualization of trained R-Tran's target embeddings over bird species (SatBird).}
        \label{fig:label_embeddings}
\end{figure}

For target embeddings $T$ in R-Tran, we train an embedding layer of all possible classes \(\{c_1, c_2, \ldots, c_m\} \) that can exist in a target $y$. In our experiments, we used textual species names from our datasets to train embeddings. Figure \ref{fig:label_embeddings} shows embeddings of R-Tran after training the model on SatBird. A 2D TSNE Visualization shows clear discrimination between bird families. In state embeddings $S$, which are essential to mark class labels as known or unknown in a feedforward, each class is given a state $s_i$, where $s_i$ takes a value of $-1$ if a class label is unknown, $0$ if known to be absent, or a positive value ($0.25, 0.5, 0.75, 1.0$) if the true probability $y$ of a class $> 0$. This $4$-bin quantization is an adaptation to the regression problem, instead of using a single value of $1$ for all present classes. This is a hyper-parameter that can be further finetuned depending on the distribution of targets, but we found $4$ to be performing best in our experiments. During inference, the model is still flexible to use information about presence $1$ or absence $0$ of species, rather than the exact encounter rate.

FeedbackProp uses the already-trained ResNet18 model to do inference with partial labels. It computes partial loss only with respect to  a set of known labels \textit{A} for input sample $I$. Then, it back-propagates this partially observed loss through the model, and iteratively update the input $I$ in order to re-compute the predictions on the set of unknown classes \textit{B}.

\subsection{Evaluation}
We report metrics regression metrics, MSE and MAE. Furthermore, we report the top-$10$, top-$30$ and top-$k$ accuracies, representing the number of species present in the top $k$ predicted species and the top $k$ observed species. In Tables \ref{songbirdvs.nonsongbird} and \ref{satbird_satbutterfly}, all metrics reported are masked for the set of classes indicated whether it is set $A$ or set $B$, using a binary mask.

\end{document}